\pdfoutput=1

\documentclass[11pt]{article}

\usepackage{emnlp2021}

\usepackage{times}
\usepackage{latexsym}
\usepackage{xspace}

\usepackage[T1]{fontenc}

\usepackage[utf8]{inputenc}
\usepackage{multirow}
\usepackage{arydshln}
\usepackage{microtype}

\usepackage{tabularx, booktabs, makecell, caption}
\usepackage{siunitx}
\usepackage{subcaption}
\usepackage{wrapfig}
\usepackage{makecell}
\usepackage{graphicx}

\newcommand{\bert}{\textsc{Bert}\xspace}
\newcommand{\roberta}{\textsc{RoBERTa}\xspace}
\newcommand{\electra}{\textsc{Electra}\xspace}
\newcommand{\reformer}{\textsc{Reformer}\xspace}
\newcommand{\longformer}{\textsc{LongFormer}\xspace}
\newcommand{\legalbert}{\textsc{Legal-Bert}\xspace}
\newcommand{\biobert}{\textsc{BioBert}\xspace}
\newcommand{\clinicalbert}{\textsc{Clinical-Bert}\xspace}
\newcommand{\lawformer}{\textsc{LawFormer}\xspace}
\newcommand{\name}{\textsc{LegalRelectra}\xspace}

\usepackage[T1]{fontenc}

\usepackage[utf8]{inputenc}

\usepackage{microtype}

\title{\name: Mixed-domain Language Modeling \\for Long-range Legal Text Comprehension}

\author{
Wenyue Hua$^{1,2}$, Yuchen Zhang$^{3}$\thanks{$^\ast$Work done while intern at Claudius Legal Intelligence.},  Zhe Chen$^{1,4}$, Josie Li$^{5}$\footnotemark[1], and Melanie Weber$^{1,6}$ \\
 $^1$Claudius Legal Intelligence, \\
 $^2$Rutgers University, $^3$University of Pennsylvania, $^4$Princeton University, \\ $^5$University of California, San Diego, $^6$Harvard University \\
 \texttt{\{wenyue,zhe,melanie\}@claudius.ai}\\ \texttt{zycalice@seas.upenn.edu}, \texttt{jol027@ucsd.edu}
 }

\begin{document}
\maketitle

\begin{abstract}
The application of Natural Language Processing (NLP) to specialized domains, such as the law, has recently received a surge of interest. As many legal services rely on processing and analyzing large collections of documents, automating such tasks with NLP tools emerges as a key challenge. Many popular language models, such as \bert\cite{bert} or \roberta\cite{roberta}, are general-purpose models, which 
have limitations on processing specialized legal terminology and syntax. In addition, legal documents may contain specialized vocabulary from other domains, such as medical terminology in personal injury text. Here, we propose \name, a legal-domain language model that is trained on mixed-domain legal and medical corpora. We show that our model improves over general-domain and single-domain medical and legal language models when processing mixed-domain (personal injury) text. Our training architecture implements the \electra framework, but utilizes \reformer instead of \bert for its generator and discriminator. We show that this improves the model's performance on processing long passages and results in better long-range text comprehension.
\end{abstract}

\section{Introduction}

Following the striking success of large language models, the development of specialized language models that are adapted to domain-specific syntax and vocabulary, has received increasing attention. In many application domains, such as medicine or the law, raw data is often given in text format, creating a need for NLP tools that aid in automating data processing. 

General-domain models typically lack expressivity on specialized domains. Rare words and domain-specific meanings of vocabulary are difficult to process with general-domain models. Thus, domain adaption had to be addressed in downstream tasks previously. In contrast, recent literature proposes to integrate domain adaptation into the pre-training stage. Such domain-specific pre-trained models have been developed for a range of domains, including the law~(\legalbert~\cite{legalbert}), medicine~(\clinicalbert~\cite{clinicalbert}), biomedical sciences~( \biobert~\cite{biobert}) and finance~\textsc{FinBert}~\citep{yang2020finbert}, among others. 

In this work we focus on pre-trained models for applications in the law with specialization to personal injury civil suits. The analysis of legal proceedings relies on access to case data, which is often given in the form of legal documents. Processing such documents presents a challenge for general-purpose language models, due to the specialized terminology and syntax conventions in the law. A natural remedy is the development of a specialized legal-domain language model that is trained on legal text~\citep{legalbert,lawformer}. There are two key challenges in developing legal-domain models. The first is the model's capability to process long passages. 
 Extracting key legal information, such as the plaintiff and defendant in a case, requires long-range text comprehension. However, most legal texts are much longer than the 512 tokens, the typical limit for \bert-based models.
 In this paper, we describe an architecture that increases the maximum passage length to 8,092 tokens. A second challenge arises from  specialized terminology from other domains in legal text. Here, we consider the example of personal injury case data, which often contains medical terminology, such as descriptions of diagnoses and treatments. In such cases, we require a language model that can process specialized text from more than one domain. Thus we train on a mixed legal and medical domain corpus.

\subsection{Summary of contributions}
The main contributions of this paper are as follows: 
\begin{enumerate}
    \vspace{-0.5em}
    \item We describe a novel model architecture (\textsc{Relectra}) that adapts the popular \electra model to the processing of long passages. For this, we replace the \bert generators and discriminators with \reformer.
    \vspace{-0.5em}
    \item We describe a training procedure for mixed-domain language models that are adapted to processing text from more than one domain. 
    \vspace{-0.5em}
    \item We investigate and demonstrate the benefits of training a domain-specific tokenizer as opposed to pre-training with the standard \bert tokenizer.
\end{enumerate}
The resulting model, \name, is well equipped to process long passages of mixed-domain text (here, personal injury cases, i.e., mixed legal and medical domain). We demonstrate the benefits of our framework in a range of benchmark experiments (Named Entity Recognition) against state of the art general and domain-specific models.

\section{Related Work}
Recently, there has been growing interest in utilizing Machine Learning in the legal domain (Legal Artificial Intelligence), including for judgement prediction~\citep{chalkidis2019neural,medvedeva2020using}, the analysis of fairness in legal proceedings~\citep{kleinberg,justfair,avery2020technology,sargent2021identifying}, as well as legal document analysis~\citep{zhong-etal-2020-nlp,grover-etal-2003-summarising,sulea2017exploring}. The development of specialized legal language models can aid in the latter.
Many variations of transformers that are adapted to specialized domains have been proposed in the literature (e.g.,~\citep{clinicalbert,legalbert,yang2020finbert, lawformer, medbert}) and demonstrated to be more efficient and accurate than \bert on their specialized domains. However, most of these pre-trained domain-specific models do not adopt new frameworks, but rely on the classical \bert architecture. To the best of our knowledge, only 
~\lawformer~\cite{lawformer}, which is a Chinese legal domain pre-trained model, utilizes~\longformer~\citep{longformer}, instead of~\bert. This architecture change allows~\lawformer to process longer passages (up to 4,096 tokens). Here, we adopt the \electra~\cite{electra} framework, which has been shown to be more data- and parameter-efficient than \bert.
In addition, our model utilizes \reformer as generator and discriminator, which significantly improves over the maximum text length of~\longformer (up to 8,092 tokens). 
In this paper, we show that the resulting model (\name) outperforms \bert, as well as state of the art single-domain adapted models on a downstream task (Named Entity Recognition, sec.~\ref{exp:ner}).

Moreover, some current state of the art domain-specific language models are pretrained using a general-domain tokenizer (e.g., the \bert tokenizer in \legalbert) to preprocess the input data. Here, we train a \emph{domain-specific} tokenizer to pre-train \name.
Our results indicate that this improves the training process and downstream task (sec.~\ref{exp:tokenizer} and~\ref{exp:pretrain}). 

Lastly, we note that all current domain-specific models focuses on adaption to a \emph{single} domain. In contrast, our model is trained on both legal and medical domains motivated by applications in processing personal injury text. Our experimental results demonstrate that \name performs competitively against general-purpose and single-domain models (sec.~\ref{exp:ner}).

\section{Language Modeling for Legal Text}
\begin{table*}[ht]
\begin{tabular}{lcl}
      \toprule
      \textbf{Data Set} & \textbf{Size} & \textbf{Description}  \\
      \midrule
      Legal & 6G & Text excerpts from US case law. \\
      Medical & 3G & Doctor's notes and letters from \textsc{MIMIC} and \textsc{MIMIC-CXR} data bases.
 \\
      Mixed & 3G & \makecell[l]{Personal injury case descriptions from Supreme Court opinions, academic \\ literature, \textsc{CourtListener}, \textsc{BYU Law}, case descriptions from attorneys.}
 \\
       \bottomrule
    \end{tabular}
    \caption{Training data, consisting of a 12GB corpus consisting of legal text, medical text, mixed-domain text with legal and medical terminology, as well as general English text. }
    \label{tab:training-data}
  \end{table*}
%
We utilize the \electra framework as basic model architecture. However, in contrast to the classical \electra structure, we replace the \bert generator and discriminator with \reformer models. In this section, we recall the structure of \electra and \reformer.

\textbf{\electra}~\citep{electra} is a sample-efficient model, which learns from all input tokens instead of just the small masked-out subset (as in \bert). It is shown to excel on the GLUE benchmark and multiple downstream NLP tasks, such as question answering. \electra consists of two sub-models, a generator and a discriminator. Given texts with 15$\%$ of the tokens masked, the generator is trained to generate the original non-masked text. Then, given the generated text, the discriminator is trained to decide whether any generated token is identical to the original token. Thus, the discriminator loss is calculated over all input tokens as it performs prediction on each token.

In order to be able to process long passages, we leverage \textbf{\reformer}~\citep{reformer}, which can process text length up to 8,192. This is in stark contrast to \bert, which can only process up to 512 tokens. Traditional transformer models incur computational and memory cost of $O(L^2)$ when computing full attention over a text of length $L$, creating a significant computational bottleneck.
However, computing full attention is unnecessary: The weighted average of attention weights and values involves ${\rm softmax}(QK^T)$, which is dominated by the largest elements in a sparse matrix. Thus for each query $q$, the model only needs to pay attention to the keys $k$ that are closest to $q$. In contrast, the \reformer model utilizes locality-sensitive-hashing (LSH) to reduce the complexity of attending over long sequences. LSH is an efficient approach for approximate nearest neighbors search in high dimensions. When using LSH to hash the $Q$ and $K$ matrices, similar $q$ and $k$ vectors are divided into the same buckets. Then standard attention is only computed for the $q$ and $k$ vectors within the same hash buckets. In our model, the generator and discriminator are two \reformer models instead of \bert models as in the original \electra framework. We name this new model \textsc{Relectra}.


\subsection{Pre-Training}
Our domain-specific language model is trained for processing personal injury legal text. \emph{Personal injury} is a legal term for an injury to the body, mind, emotions or reputation, as opposed to injury to property rights\footnote{adopted from~\url{https://www.law.cornell.edu/wex/personal_injury}}. Personal injury lawsuits 
are filed against the person or entity that caused the harm through negligence, gross negligence, reckless conduct, or intentional misconduct. Jurisdictions typically describe the injured person's medical bills, pain and suffering, as well as diminished quality of life. Thus, personal injury legal text involves both legal and medical terminology, which creates a necessity to include both legal and medical texts in the pre-training corpus.

Our legal-domain pre-training corpus contains text collected from public data bases (including \textsc{CourtListener} from the FreeLaw project~\cite{courtlistener}), as well as anonymized descriptions of civil cases. Preprocessing steps involve standard steps to remove specical characters, foreign language texts and headers.
Ideally, the model should use a large collection of personal injury texts as the pre-training corpus. However, most personal injury texts are licensed to law firms or governments making it difficult to collect a sufficiently large amount of personal injury text. Hence, we supplement our core personal injury text corpus with text from other legal branches, as well as with (pure) medical text.
Details on the sources of our training data can be found in Tab.~\ref{tab:training-data}.
It collects text from medical (3GB), legal (6GB) and mixed legal-medical (3GB) domains. The final corpus contains 12GB of text.

\begin{table*}[ht]\centering
\small
\newcolumntype{C}{>{\centering\arraybackslash}X}
\begin{tabular}{c*{9}{S}}
\toprule
\addlinespace 
&\multicolumn{4}{c}{\bert tokenizer}
&\multicolumn{4}{c}{Custom tokenizer} \\
 \cmidrule(lr){2-2}
 \cmidrule(lr){3-5}
 \cmidrule(lr){6-6}
 \cmidrule(lr){7-9}
{}&{words}&{total errors}&{legal}&{medical}&{words}&{total errors}&{legal}&{medical}
\tabularnewline
\cmidrule[\lightrulewidth](lr){1-9}\addlinespace[1ex]
$\#$ 0 & 153 & 5 & 0 & 1 & 139 & 0 & 0 & 0 \tabularnewline
$\#$ 1 & 411 & 3 & 0 & 2 & 417 & 3 & 0 & 1 \tabularnewline
$\#$ 2 & 285 & 12 & 5 & 2 & 277 & 7 & 1 & 0 \tabularnewline
$\#$ 3 & 418 & 9 & 0 & 7 & 412 & 6 & 0 & 2 \tabularnewline
$\#$ 4 & 313 & 9 & 3 & 2 & 289 & 4 & 0 & 1 \tabularnewline
$\#$ 5 & 405 & 9 & 1 & 3 & 385 & 8 & 0 & 1 \tabularnewline
$\#$ 6 & 216 & 9 & 3 & 4 & 210 & 4 & 0 & 3 \tabularnewline
$\#$ 7 & 560 & 12 & 5 & 4 & 539 & 11 & 0 & 7 \tabularnewline
$\#$ 8 & 400 & 13 & 4 & 3 & 407 & 7 & 1 & 0 \tabularnewline
$\#$ 9 & 340 & 12 & 1 & 5 & 323 & 7 & 0 & 1 \tabularnewline
\addlinespace
\bottomrule
\end{tabular}
\captionof{table}{\textbf{Tokenizer:} Evaluation of custom \name tokenizer against \bert tokenzier for ten text segments of personal injury case descriptions. We report the number of words recognized by the tokenizers, as well as the number of unique errors. In addition to the total number of errors, we report the number of errors for medical and legal terminology separately.}
\label{tab:tokenizer}
\end{table*}

\subsection{Named Entity Recognition}
\label{sec:annotation}
We validate \name by training legal and mixed-domain NER models. Training is performed using automatically (in-house) annotated legal text. In addition, we benchmark \name on general and medical domain NER tasks, for which we used publicly available annotated training data conll2003 ~\cite{tjong-kim-sang-de-meulder-2003-introduction} (general domain), MIMIC III~\cite{Johnson2016MIMICIIIAF} (medical domain)).

We chose the following labels, which are representative of an NER task that one may encounter in practise: \emph{plaintiff}, \emph{defendant} and \emph{case type}. Here, \emph{case type} categorizes the branch of civil law, which applies to the case. We distinguish among \emph{motorvehicle accidents}, \emph{slip-and-fall accidents} and \emph{work-related cases}, such as illegal termination of a work contract and negligence by a professional. For the latter, the NER task consists of identifying words that help to determine the type of a case. Case types are usually not explicitly mentioned in-text; hence, it can be challenging to categorize civil legal documents. 
The manual annotation of the training and testing data was performed as follows: 
For our legal text sources, we have access to ground truth plaintiff and defendant information from case headers. This allows us to annotate relevant phrases in the text via string matching. For the case type annotations, we create a word list for each case type, which contains frequently occurring phrases that are indicative of the respective case type. We then enrich the lists with synonyms of these phrases. After creating the initial word lists, we removed ambiguous phrases or phrases that appeared in the wordlists of more than one case type. Annotations were done via inexact string matching at first to reduce the workload for annotators, where phrases that were similar to an entry in a word list were labeled with the respective case type. After performing automatic annotations based on header information and case type word list, all cases in training, validation and testing are manually checked by three expert annotators.

\section{Experimental Setup}
\begin{table*}[ht]
\resizebox{15cm}{!}{
\begin{tabular}{ll}
      \toprule
      \textbf{Medical} & Tokenization result  \\
      \midrule
      text & {gastrointestinal complaints, neurologic changes, rashes, palpitations, orthopnea} \\
      \bert (uncased) & {gas, tro, int, estinal, complaints, ne, uro, logic, changes, rash, es, pal, pit, ations, or, th, op, nea} \\
      custom (uncased) & {gastrointestinal, complaints, neurologic, changes, rashes, palpitations, orthopnea}\\
       \midrule
      \textbf{Legal} & Tokenization result  \\
       \midrule
       text & {the nature of adjudications upon which erroneous subsequent proceedings rest}\\
       \bert (uncased) & {the, nature, of, ad, ju, dication, s, upon, which, er, rone, ous, subsequent, proceedings, rest}\\
       custom (uncased) & {the, nature, of, adjudications, upon, which, erroneous, subsequent, proceedings, rest}\\
       \bottomrule
    \end{tabular}
    }
    \caption{Tokenization example for the \bert tokenizer and our custom tokenizer.}
    \label{tab:tokenization-data}
  \end{table*}

\subsection{Tokenizer}
\emph{Tokenization} refers to the process of splitting a stream of characters into words~\cite{Grefenstette1994WhatIA}.
While often seen as part of preprocessing, good tokenization is a crucial prerequisite for good downstream performance. Tokenization is particularly important for processing domain-specific text that contains a large amount of specialized terminology. In contrast to most of the published domain-specific language models, which use the standard \bert tokenizer, we train a custom tokenizer. Our tokenizer is trained via standard Byte-Pair Encoding~\cite{subword2016}, which replaces the most common pair of consecutive bytes of data with a byte that does not occur in that data until the vocabulary size is reached.
Since our pre-training data set contains text with specialized terminology, our custom tokenizer generates more sensible tokenization of domain-specific text. An example is given in  
Tab.~\ref{tab:tokenization-data}.

\subsection{Pre-Training}
We pre-train \name on the collected legal, medical and mixed-domain text corpora described above, in batches of four samples each. Our combined corpora contain up to 30,522 sentence piece tokens. There are 120k training steps and 20k warm-up steps (linear warm-up with linear decay scheduler). We use the \textsc{AdamW} optimizer with learning rate of 1e-5 for the first 80k training steps and 1e-6 for the remaining 40k training steps. \name was trained on one NVIDIA P100 GPU and 16 GB memory, for a total of 16 days, 6 hours and 30 minutes.

\subsection{Named Entity Recognition}
We evaluate \name on a downstream task, \emph{Named Entity Recognition} (NER), with three labels for the legal domain (case type, plaintiff, defendant) and four labels for the mixed medical-legal domain (case type, plaintiff, defendant, injury). We use \bert, \clinicalbert, \legalbert and \reformer as baseline models. Each NER model is trained for 10 epochs with batch size 1, using the AdamW optimizer with learning rate 3e-5.

\section{Results}
\subsection{Tokenizer}
\label{exp:tokenizer}
The performance of tokenizers is in general difficult to evaluate, as it is highly dependent on the downstream application. To the best of our knowledge, no established convention exists in the literature. Here, we employ quantitative metrics for comparing the performance of our custom \name tokenizer with that of the standard \bert tokenizer. For our evaluation, we align the output of the \bert and \name tokenizers for ten text segments with personal injury case descriptions. All text segments contain both medical and legal terminology.  
We analyze the number of recognized words and the number of total unique errors, as well as errors in medical and legal phrases. A similar validation scheme was suggested in~\cite{habert1998towards}. Errors in abbreviations are excluded from the error count. Our results (see Tab.~\ref{tab:tokenizer}) show that the custom \name tokenizer performs better at recognizing words, as evident in the lower number of words detected. We further notice that \name tokenizer has a smaller number of errors involving legal and medical terminology, suggesting a superior performance on domain-specific text. Errors that do not involve medical or legal terminology are often (non-English) names for both tokenizers.

\subsection{Pre-training}
\label{exp:pretrain}
\begin{figure}[!h]
     \centering
     \hfill
     \begin{subfigure}[b]{0.23\textwidth}
         \centering
         \includegraphics[width=\textwidth]{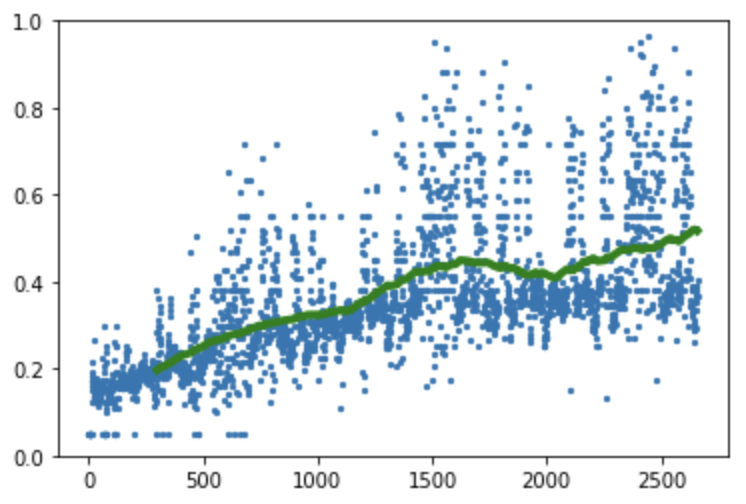}
         \caption{Generator (custom tok)}
         \label{fig:gen_acc}
     \end{subfigure}
     \hfill
     \begin{subfigure}[b]{0.23\textwidth}
         \centering
         \includegraphics[width=\textwidth]{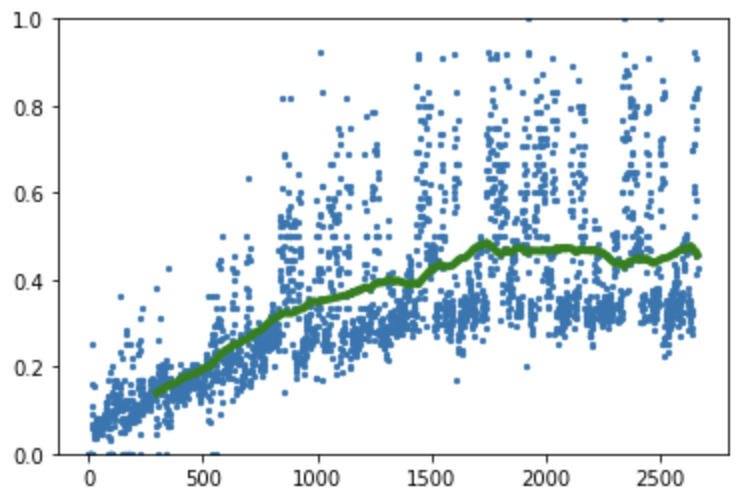}
         \caption{Generator (\bert tok)}
         \label{fig:bert_gen_acc}
     \end{subfigure}
     \hfill \\
     \hfill
     \begin{subfigure}[b]{0.23\textwidth}
         \centering
         \includegraphics[width=\textwidth]{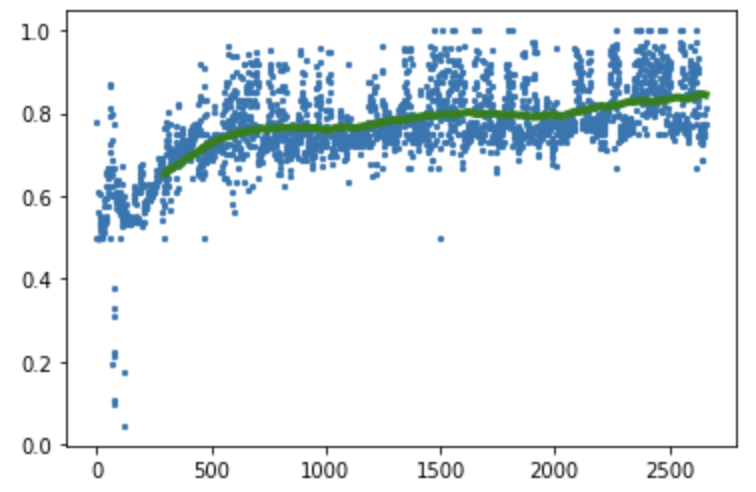}
         \caption{Discriminator (custom)}
         \label{fig:dis_acc}
     \end{subfigure}
     \hfill
     \begin{subfigure}[b]{0.23\textwidth}
         \centering
         \includegraphics[width=\textwidth]{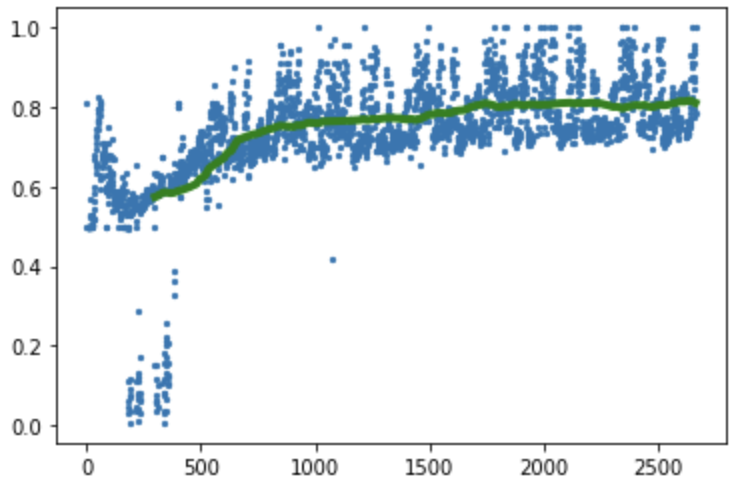}
         \caption{Discriminator (\bert)}
         \label{fig:bert_dis_acc}
     \end{subfigure}
        \caption{Masked Language Modeling (MLM) accuracy of generator and discriminator of \name with our custom tokenizer and the standard \bert tokenizer. Here, one dot represents the accuracy score of one datapoint (evaluated every 100 datapoints). The line is a smoothed function of averaged accuracy scores over intervals of length 200 aggregating the scores of the 100 data points before and after respectively.}
        \label{fig:pretrain}
\end{figure}
\begin{figure}[!ht]
    \centering
    \includegraphics[scale=0.4]{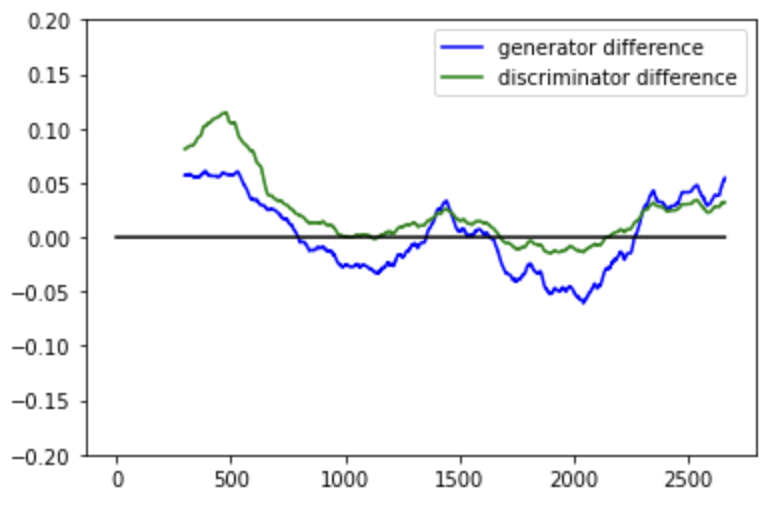}
    \caption{Differences of Generator and Discriminator accuracy between \name models pretrained with custom tokenizer and \bert tokenizer.}
    \label{fig:diff}
\end{figure}

We evaluate \name trained with our custom (domain-specific) tokenizer in comparison with a second \name model that was trained using the standard (general-domain) \bert tokenizer. We report both generator and discriminator accuracy after 120k training steps. The pretraining task for the generator is Masked language modeling (MLM). In MLM,  the input is corrupted by replacing some tokens with ``[MASK]''. Then we train a model to reconstruct the original tokens. The pretraining task for the discriminator is to predict whether each token in the corrupted input was replaced by a generator sample or not. The results for both models are shown in Fig.~\ref{fig:pretrain}. We observe that the pretraining performance of the generator and discriminator is comparable. Notably, both the generator and discriminator model trained with our custom tokenizer improve over the model trained with the \bert tokenizer towards the end of the training process. Fig.~\ref{fig:diff} shows the difference in generator and discriminator accuracy between both models. 

\subsection{Named Entity Recognition}
\label{exp:ner}
To evaluate \name on a downstream task, we analyze the performance of legal domain and mixed legal-medical domain NER models trained with \name in comparison with benchmark NER models trained with \bert, \legalbert, \clinicalbert, \reformer and \name with \bert tokenizer. For experiments on \bert, \legalbert, \clinicalbert which have a maximum token length of 512, we stride and chunk the training and test passage. For experiments on \reformer, \name and \name with \bert tokenizer, we use a maximum token length of 1536 evaluated on validation data. 

\begin{table*}[ht] \centering
\setlength{\tabcolsep}{5pt}
\footnotesize
\newcolumntype{C}{>{\centering\arraybackslash}X}
\begin{tabular}{l*{11}{S}}
\toprule
&\multicolumn{3}{c}{precision} 
&\multicolumn{3}{c}{recall} 
&\multicolumn{3}{c}{f1}
&\multicolumn{1}{c}{f1} \\
 \cmidrule(lr){2-4}
 \cmidrule(lr){5-7}
 \cmidrule(lr){8-10}
 \cmidrule(lr){11-11}
{Legal domain}&{TYPE}&{PLT}&{DEF}&{TYPE}&{PLT}&{DEF}&{TYPE}&{PLT}&{DEF}&{all}
\tabularnewline
\cmidrule[\lightrulewidth](lr){1-11}
\bert & 95.00 & 82.75 & 82.86 & 76.00 & 64.86 & 87.88 & 84.44 & 72.73 & 85.29 & 80.45 \\
\clinicalbert & 95.00 & 86.95 & 87.88 & 79.17 & 58.89 & 90.63 & 86.36 & 70.17 & 89.23 & 81.93 \\
\legalbert & 84.70 & 82.31 & 88.56 & 77.11 & 65.34 & 80.59 & 80.73 & 72.85 & 84.39 & 79.32 \\
\reformer & 87.51 & 73.69 & 92.33 & 76.25 & 87.97 & 89.31 & 81.48 & 80.20 &  90.79 & 84.16 \\
\name & 89.66 & 84.31 & 100.00 & 74.28 & 84.31 & 83.33 & 81.25 & 84.31 &  90.91 & \textbf{85.93} \\
\makecell[l]{\name \\(\bert tok)} & 95.23 & 85.37 & 97.56 & 55.56 & 67.31 & 86.96 & 70.17 & 75.27 & 91.95 & 80.12\\
\bottomrule
\end{tabular}
\captionof{table}{\textbf{NER-legal:} Performance of \name on legal text in comparison with \bert, \clinicalbert, \legalbert and \name (\bert tokenizer) for case type (TYPE), defendant (DEF) and plaintiff (PLT). \label{tab:legal}}
\end{table*}
\begin{table*}[ht] \centering
\setlength{\tabcolsep}{4pt}
\footnotesize
\newcolumntype{C}{>{\centering\arraybackslash}X}
\begin{tabular}{l*{14}{S}}
\toprule
\addlinespace 
&\multicolumn{4}{c}{precision} 
&\multicolumn{4}{c}{recall} 
&\multicolumn{4}{c}{f1}
&\multicolumn{1}{c}{f1} \\
 \cmidrule(lr){2-5}
 \cmidrule(lr){6-9}
 \cmidrule(lr){10-13}
 \cmidrule(lr){14-14}
{Mixed domain}&{TYPE}&{PLT}&{DEF}&{PROB}&{TYPE}&{PLT}&{DEF}&{PROB}&{TYPE}&{PLT}&{DEF}&{PROB}&{all}\\
\cmidrule[\lightrulewidth](lr){1-14}
\bert & 70.59 & 82.76 & 71.43 & 82.61 & 57.14 & 64.86 & 80.65 & 76.00 & 63.15 & 72.72 & 75.76 & 79.17 & 73.39 \\
\clinicalbert & 84.00 & 82.91 & 75.68 & 88.46 & 69.23 & 58.82 & 87.32 & 92.00 & 75.90 & 80.08 & 68.11 & 90.19 & 78.55 \\
\legalbert & 88.88 & 80.00 & 83.87 & 85.00 & 66.67 & 60.61 & 86.67 & 66.67 & 76.19 & 68.97 & 85.25 & 77.27 & 77.07 \\
\reformer & 89.92 & 83.44 & 86.95 & 83.21 & 63.56 & 62.17 & 84.38 & 78.92 & 74.48 & 71.25 & 85.65 & 81.01 & 78.10 \\
\name & 91.30 & 85.71 & 95.12 & 86.96 & 58.33 & 57.69 & 92.86 & 76.92 & 71.18 & 68.97 & 93.98 & 81.63 & \textbf{78.57}\\
\makecell[l]{\name \\(\bert tok)} & 90.47 & 81.58 & 95.00 & 73.91 & 54.28 & 59.62 & 79.17 & 65.38 & 67.86 & 68.89 & 86.36 & 69.39 & 74.20 \\
\bottomrule
\end{tabular}
\captionof{table}{\textbf{NER-mixed:} Performance of \name in mixed domain with labels case type (TYPE), defendant (DEF) plaintiff (PLT), medical problem (PROB). \label{tab:mixed}}
\end{table*}
The results are given in Tables~\ref{tab:legal} (legal domain) and~\ref{tab:mixed} (mixed domain). We observe that \name outperforms both the general-domain \bert, as well as the specialized \legalbert and \clinicalbert models on both domains (with respect to the overall f1 score). It also performs better than \reformer which is pretrained on the same corpus, demonstrating the benefits of \electra training framework.   Notably, both \name and \reformer trained with our custom tokenizer outperform \name trained with the \bert tokenizer, which demonstrates the benefits of a domain-specific tokenizer. 

\subsection{Result Analysis} 
A key observation in our experimental results is the superior performance of \name in comparison with \legalbert.
We notice that, surprisingly, \legalbert does not perform as well as expected on legal case terminology, specifically the recognition of plaintiff and defendant. For example, it may confuse other entities, such as the defendant's attorney, as defendant or plaintiff. While the extracted information is relevant to the case, it is not precise enough to be useful for further downstream analysis. 
Secondly, \legalbert does not perform well on case type recognition in legal text (\name does not perform well on case type recognition in mixed text either, see below). Thirdly, \legalbert misses medical terminology, as it is not pretrained on medical data. Examples include TMJ (an abbreviation for for Temporomandibular joint), Myofascial Pain Syndrome and psychological injuries such as emotional distress. \clinicalbert does not perform well on legal case terminology, such as plaintiff and defendant recognition. This is expected, as it is not pretrained on legal text. However, it indeed outperforms all other models on medical terminology recognition, including \name, which is also pre-trained on some medical data. It is unexpectedly good at the recognition of case types. \bert sometimes misses plaintiff or defendant throughout the whole case text, indicating its limited ability to recognize legal terminology. 

In summary, all three models show a limited ability to recognize legal terminology in \emph{long} texts. Errors arise due to the limited number of tokens that the models can process. In some texts the plaintiff and defendant information is only given in the beginning of the text and can therefore not be obtained from partial text segments. In contrast, \name's ability to process larger text segments results in a superior performance in plaintiff and defendant recognition. Its performance on medical problem recognition is better than all other models except \clinicalbert, indicating that the mixed pre-trained data does help on both legal and medical feature learning. \reformer performs slightly worse on precision but better on recall, but an overall worse result on F1, indicating that the \electra training framework indeed improves model performance.
Comparing \name trained with our custom tokenizer and the standard \bert tokenizer, we notice that the performance on legal feature recognition is comparable. However, the recognition of medical terminology is negatively affected by the \bert tokenizer. This indicates that wrong tokenization of medical terminology affects the model performance more negatively than wrong tokenization of legal terminology.



\section{Conclusions}
In this paper, we introduced \name, a language model that is specialized to process mixed legal and medical domain (personal injury) text. This was achieved by pre-training with a corpus consisting of legal, medical and mixed domain (personal injury) text. We demonstrate in validation experiments that \name outperforms general-purpose language models (e.g., \bert), as well as specialized legal-domain models (e.g., \legalbert) on legal and mixed-domain NER. As technical contributions, we proposed a novel model architecture that allows for improved performance on long-range text comprehension.

\name provides a pretrained model for personal injury text with a special focus on enabling long-range text comprehension. It lends itself to a plethora of applications that involve legal case documents, including summarization or extraction of key information for civil suits, identifying patterns and trends in legal proceedings, identifying precedent in past cases, among others. In addition, legal language models may aid in summarizing and analysing legal scholarship. 

We demonstrate the applicability of \name on a downstream task, for which we train a Name Entity Recognition (NER) model. In practise, such an NER model may be applied to extract key legal information from case documents, such as the identities of plaintiff and defendant, medical injuries and civil case type. With that, NER enables a simple summarization of civil suits, which may serve as a basis for further case analytics.

\section{Limitations}
\label{sec:limitations}
There are several limitations in our training and validation setup, addressing of which may lead to significant improvements. As discussed above, the ideal pre-training corpus for a personal injury language model would consist of large collections of personal injury text. However, due to the restricted access to such data, it is difficult to collect a sufficiently large text corpus. Thus, we supplement our pre-training data with text from other legal branches and (pure) medical text, which may have decreased the model's performance. 

Further limitations arise in the performance evaluation presented here. Testing and validation against benchmarks could have been more extensive, for instance by evaluating \name on a second downstream task in addition to NER. Moreover, the downstream training data annotation (see sec.~\ref{sec:annotation}) was partially automated. Instead of string matching, we could have annotated the NER training data manually, which would have been more accurate.

\section{Acknowledgements}
This work was supported by NSF Grant 2112315. 
We are pleased to acknowledge that the reported work was substantially performed using the Princeton Research Computing resources at Princeton University which is a consortium of groups led by the Princeton Institute for Computational Science and Engineering (PICSciE) and Office of Information Technology's Research Computing.

\bibliography{anthology,custom}

\begin{thebibliography}{26}
\expandafter\ifx\csname natexlab\endcsname\relax\def\natexlab#1{#1}\fi

\bibitem[{Avery and Cooper(2020)}]{avery2020technology}
Joseph Avery and Joel Cooper. 2020.
\newblock Technology in the legal system.
\newblock \emph{Bias in the Law: A Definitive Look at Racial Prejudice in the
  US Criminal Justice System}, 161.

\bibitem[{Beltagy et~al.(2020)Beltagy, Peters, and Cohan}]{longformer}
Iz~Beltagy, Matthew~E Peters, and Arman Cohan. 2020.
\newblock Longformer: The long-document transformer.

\bibitem[{Chalkidis et~al.(2019)Chalkidis, Androutsopoulos, and
  Aletras}]{chalkidis2019neural}
Ilias Chalkidis, Ion Androutsopoulos, and Nikolaos Aletras. 2019.
\newblock Neural legal judgment prediction in english.
\newblock \emph{arXiv preprint arXiv:1906.02059}.

\bibitem[{Chalkidis et~al.(2020)Chalkidis, Fergadiotis, Malakasiotis, Aletras,
  and Androutsopoulos}]{legalbert}
Ilias Chalkidis, Manos Fergadiotis, Prodromos Malakasiotis, Nikolaos Aletras,
  and Ion Androutsopoulos. 2020.
\newblock Legal-bert: The muppets straight out of law school.

\bibitem[{Ciocanel et~al.(2020)Ciocanel, Topaz, Santorella, Sen, Smith, and
  Hufstetler}]{justfair}
Maria-Veronica Ciocanel, Chad~M. Topaz, Rebecca Santorella, Shilad Sen,
  Christian~Michael Smith, and Adam Hufstetler. 2020.
\newblock \href {https://doi.org/10.1371/journal.pone.0241381} {Justfair:
  Judicial system transparency through federal archive inferred records}.
\newblock \emph{PLOS ONE}, 15(10):1--20.

\bibitem[{Clark et~al.(2019)Clark, Luong, Le, and Manning}]{electra}
Kevin Clark, Minh-Thang Luong, Quoc~V Le, and Christopher~D Manning. 2019.
\newblock Electra: Pre-training text encoders as discriminators rather than
  generators.
\newblock In \emph{International Conference on Learning Representations}.

\bibitem[{Grefenstette and Tapanainen(1994)}]{Grefenstette1994WhatIA}
Gregory Grefenstette and Pasi Tapanainen. 1994.
\newblock What is a word, what is a sentence? problems of tokenization.

\bibitem[{Grover et~al.(2003)Grover, Hachey, and
  Korycinski}]{grover-etal-2003-summarising}
Claire Grover, Ben Hachey, and Chris Korycinski. 2003.
\newblock \href {https://aclanthology.org/W03-0505} {Summarising legal texts:
  Sentential tense and argumentative roles}.
\newblock In \emph{Proceedings of the {HLT}-{NAACL} 03 Text Summarization
  Workshop}, pages 33--40.

\bibitem[{Habert et~al.(1998)Habert, Adda, Adda-Decker, de~Mar{\"e}uil,
  Ferrari, Ferret, Illouz, and Paroubek}]{habert1998towards}
Benoit Habert, Gilles Adda, Martine Adda-Decker, P~Boula de~Mar{\"e}uil, Serge
  Ferrari, Olivier Ferret, Gabriel Illouz, and Patrick Paroubek. 1998.
\newblock Towards tokenization evaluation.
\newblock In \emph{Proceedings of LREC}, volume~98, pages 427--431.

\bibitem[{Huang et~al.(2019)Huang, Altosaar, and Ranganath}]{clinicalbert}
Kexin Huang, Jaan Altosaar, and Rajesh Ranganath. 2019.
\newblock Clinicalbert: Modeling clinical notes and predicting hospital
  readmission.

\bibitem[{Johnson et~al.(2016)Johnson, Pollard, Shen, wei H.~Lehman, Feng,
  Ghassemi, Moody, Szolovits, Celi, and Mark}]{Johnson2016MIMICIIIAF}
Alistair E.~W. Johnson, Tom~J. Pollard, Lu~Shen, Li~wei H.~Lehman, Mengling
  Feng, Mohammad~Mahdi Ghassemi, Benjamin Moody, Peter Szolovits, Leo~Anthony
  Celi, and Roger~G. Mark. 2016.
\newblock Mimic-iii, a freely accessible critical care database.
\newblock \emph{Scientific Data}, 3.

\bibitem[{Kenton and Toutanova(2019)}]{bert}
Jacob Devlin Ming-Wei~Chang Kenton and Lee~Kristina Toutanova. 2019.
\newblock Bert: Pre-training of deep bidirectional transformers for language
  understanding.
\newblock In \emph{Proceedings of NAACL-HLT}, pages 4171--4186.

\bibitem[{Kitaev et~al.(2019)Kitaev, Kaiser, and Levskaya}]{reformer}
Nikita Kitaev, Lukasz Kaiser, and Anselm Levskaya. 2019.
\newblock Reformer: The efficient transformer.
\newblock In \emph{International Conference on Learning Representations}.

\bibitem[{Kleinberg et~al.(2020)Kleinberg, Ludwig, Mullainathan, and
  Sunstein}]{kleinberg}
Jon Kleinberg, Jens Ludwig, Sendhil Mullainathan, and Cass~R. Sunstein. 2020.
\newblock Algorithms as discrimination detectors.
\newblock \emph{Proceedings of the National Academy of Sciences},
  117(48):30096--30100.

\bibitem[{Lee et~al.(2020)Lee, Yoon, Kim, Kim, Kim, So, and Kang}]{biobert}
Jinhyuk Lee, Wonjin Yoon, Sungdong Kim, Donghyeon Kim, Sunkyu Kim, Chan~Ho So,
  and Jaewoo Kang. 2020.
\newblock Biobert: a pre-trained biomedical language representation model for
  biomedical text mining.
\newblock volume~36, pages 1234--1240. Oxford University Press.

\bibitem[{Liu et~al.(2019)Liu, Ott, Goyal, Du, Joshi, Chen, Levy, Lewis,
  Zettlemoyer, and Stoyanov}]{roberta}
Yinhan Liu, Myle Ott, Naman Goyal, Jingfei Du, Mandar Joshi, Danqi Chen, Omer
  Levy, Mike Lewis, Luke Zettlemoyer, and Veselin Stoyanov. 2019.
\newblock Roberta: A robustly optimized bert pretraining approach.
\newblock \emph{arXiv preprint arXiv:1907.11692}.

\bibitem[{Medvedeva et~al.(2020)Medvedeva, Vols, and
  Wieling}]{medvedeva2020using}
Masha Medvedeva, Michel Vols, and Martijn Wieling. 2020.
\newblock Using machine learning to predict decisions of the european court of
  human rights.
\newblock \emph{Artificial Intelligence and Law}, 28(2):237--266.

\bibitem[{Rasmy et~al.(2021)Rasmy, Xiang, Xie, Tao, and Zhi}]{medbert}
Laila Rasmy, Yang Xiang, Ziqian Xie, Cui Tao, and Degui Zhi. 2021.
\newblock Med-bert: pretrained contextualized embeddings on large-scale
  structured electronic health records for disease prediction.
\newblock \emph{NPJ digital medicine}, 4(1):1--13.

\bibitem[{Sargent and Weber(2021)}]{sargent2021identifying}
Jackson Sargent and Melanie Weber. 2021.
\newblock Identifying biases in legal data: An algorithmic fairness
  perspective.
\newblock \emph{arXiv preprint arXiv:2109.09946}.

\bibitem[{Sennrich and Birch(2016)}]{subword2016}
Barry~Haddow Sennrich, Rico and Alexandra Birch. 2016.
\newblock Neural machine translation of rare words with subword units.
\newblock In \emph{2016 Association of Computational Linguistics}.

\bibitem[{Sulea et~al.(2017)Sulea, Zampieri, Malmasi, Vela, Dinu, and
  Van~Genabith}]{sulea2017exploring}
Octavia-Maria Sulea, Marcos Zampieri, Shervin Malmasi, Mihaela Vela, Liviu~P
  Dinu, and Josef Van~Genabith. 2017.
\newblock Exploring the use of text classification in the legal domain.
\newblock \emph{arXiv preprint arXiv:1710.09306}.

\bibitem[{{The Free Law Project}(2021)}]{courtlistener}
{The Free Law Project}. 2021.
\newblock \href {https://www.courtlistener.com} {Courtlistener}.

\bibitem[{Tjong Kim~Sang and
  De~Meulder(2003)}]{tjong-kim-sang-de-meulder-2003-introduction}
Erik~F. Tjong Kim~Sang and Fien De~Meulder. 2003.
\newblock \href {https://www.aclweb.org/anthology/W03-0419} {Introduction to
  the {C}o{NLL}-2003 shared task: Language-independent named entity
  recognition}.
\newblock In \emph{Proceedings of the Seventh Conference on Natural Language
  Learning at {HLT}-{NAACL} 2003}, pages 142--147.

\bibitem[{Xiao et~al.(2021)Xiao, Hu, Liu, Tu, and Sun}]{lawformer}
Chaojun Xiao, Xueyu Hu, Zhiyuan Liu, Cunchao Tu, and Maosong Sun. 2021.
\newblock Lawformer: A pre-trained language model for chinese legal long
  documents.
\newblock volume~2, pages 79--84. Elsevier.

\bibitem[{Yang et~al.(2020)Yang, Uy, and Huang}]{yang2020finbert}
Yi~Yang, Mark Christopher~Siy Uy, and Allen Huang. 2020.
\newblock Finbert: A pretrained language model for financial communications.
\newblock \emph{arXiv preprint arXiv:2006.08097}.

\bibitem[{Zhong et~al.(2020)Zhong, Xiao, Tu, Zhang, Liu, and
  Sun}]{zhong-etal-2020-nlp}
Haoxi Zhong, Chaojun Xiao, Cunchao Tu, Tianyang Zhang, Zhiyuan Liu, and Maosong
  Sun. 2020.
\newblock How does {NLP} benefit legal system: A summary of legal artificial
  intelligence.
\newblock In \emph{Proceedings of the 58th Annual Meeting of the Association
  for Computational Linguistics}, Online. Association for Computational
  Linguistics.

\end{thebibliography}
\bibliographystyle{acl_natbib}

\end{document}